\documentclass{article}

\usepackage{PRIMEarxiv}

\usepackage{times}

\usepackage{courier} 

\usepackage{array}

\usepackage{graphicx}    
\usepackage{epstopdf}

\usepackage{times}
\usepackage{graphicx}    
\usepackage{epstopdf}

\usepackage{wrapfig}

\usepackage{amssymb}
\usepackage{tweaklist}
\usepackage[footnotesize]{caption}
\usepackage{dsfont}

\usepackage[tiny,compact]{titlesec}

\usepackage{csquotes}

\usepackage{extarrows}
\usepackage{amsthm}

\renewcommand{\enumhook}{\setlength{\topsep}{0pt}%
	\setlength{\itemsep}{0pt}
	\setlength{\leftmargin}{1.5em}
}
\renewcommand{\itemhook}{\setlength{\topsep}{0pt}%
	\setlength{\itemsep}{0pt}
	\setlength{\leftmargin}{1.5em}
}

\usepackage{xcolor} 

\newcommand{\comment}[1]{}

\title{Enhancing Student Learning with LLM-Generated Retrieval Practice Questions: 
An Empirical Study in Data Science Courses
}

\author{
    Yuan An  \\
    College of Computer and Informatics \\
    Drexel University \\
    Philadelphia, PA, 19104\\
    \texttt{ya45@drexel.edu} \\
    \and
    \textbf{John Liu}  \\
    College of Computer and Informatics \\
    Drexel University \\
    Philadelphia, PA, 19104\\
    \texttt{jl4582@drexel.edu} \\
    \and
    \textbf{Niyam Acharya}  \\
    College of Computer and Informatics \\
    Drexel University \\
    Philadelphia, PA, 19104\\
    \texttt{nka42@drexel.edu} \\
    \and
    \textbf{Ruhma Hashmi}  \\
    College of Computer and Informatics \\
    Drexel University \\
    Philadelphia, PA, 19104\\
    \texttt{rh927@drexel.edu} \\
}

\begin{document}

\maketitle

\begin{abstract}
Retrieval practice is a well-established pedagogical technique known to significantly
enhance student learning and knowledge retention. However, generating high-quality retrieval
practice questions is often time-consuming and labor intensive for instructors,
especially in rapidly evolving technical subjects. Large Language Models (LLMs) 
offer the potential to automate this process by generating questions in response 
to prompts, yet the effectiveness of LLM-generated retrieval practice 
on student learning remains to be established.
In this study, we conducted an empirical study involving two college-level 
data science courses, with approximately 60 students.
We compared learning outcomes during one week in which students received LLM-generated multiple-choice 
retrieval practice questions to those from a week in which no such questions were provided.
Results indicate that students exposed to LLM-generated retrieval practice achieved 
significantly higher knowledge retention, with an average accuracy of 89\%, 
compared to 73\% in the week without such practice. These findings suggest 
that LLM-generated retrieval questions can effectively support 
student learning and may provide a scalable solution for integrating 
retrieval practice into real-time teaching.
However, despite these encouraging outcomes and the potential time-saving benefits, 
cautions must be taken, as the quality of LLM-generated questions can vary.
Instructors must still manually verify and revise the generated questions before 
releasing them to students.

\end{abstract}


%
%
\section{Introduction}
\label{sec:introduction}

Retrieval practice \cite{roediger2011critical}, frequently termed the ``testing effect" \cite{Agarwal2008-examining}, 
is a teaching technique that has been extensively studied. 
Empirical evidence has clearly indicated its ability 
to enhance long-term memory retention and overall 
learning \cite{Roediger2006-test-enhanced,Agarwal2008-examining,Nadel2012-memory,Roediger2011-test-enhanced,Lyle2011-retrieving,Agarwal2019-retrieval}. 
This technique involves the active recall of information from memory, a process 
that inherently strengthens neural connections and improves the future retrieval of that 
information. The historical roots of retrieval practice can be traced back to the observations of 
Francis Bacon in 1620 \cite{wikipediaTestingEffect} and William James in 1890 \cite{wikipediaTestingEffect}.
Modern experimental studies were conducted in the early 20th century through 
the pioneering work of researchers such as Abbott (1909) \cite{abbott1909recall} 
and Gates (1917) \cite{gates1917recitation}.

It was not until the 21st century, nearly 100 years later, that retrieval 
practice gained popularity among educators as a teaching strategy \cite{agarwal2024personal}.
Roediger and Karpicke \cite{Roediger2006-test-enhanced,karpicke2009metacognitive,Roediger2011-test-enhanced}
has consistently demonstrated the effectiveness of retrieval practice over alternative 
study methods, including concept mapping or repeated reading. 
Using retrieval practice proves effective across diverse age groups and subject domains, and its benefits 
extend to improvement of higher-order thinking. Furthermore, retrieval practice 
has been shown to improve metacognition. Students are better able to identify 
gaps in their knowledge and direct their study efforts more effectively.

\noindent
\textbf{Concrete Retrieval Practice Activities.} 
Retrieval practice can be integrated into classrooms through a 
range of in-class activities, such as \emph{warm-up quizzes, live polling, 
multiple-choice questions (MCQs), think–pair–share, one-minute write-ups, 
concept mapping from memory}, and \emph{brain dumps}. 
In a college-level data science programming course, for example, 
an instructor might pose the following low-stakes MCQ
that prompts students to recall and apply key programming concepts:

\noindent
\textbf{Question:}\texttt{"Which of the following Python code snippets correctly calculates 
the mean of the columns of a pandas DataFrame while ignoring missing values (NaNs)?"}

\vspace{0.5em}
\noindent
\textbf{Answers (Choose one):}
{\setlength{\parindent}{1cm} 
 \setlength{\hangindent}{1cm} 

(a) \texttt{data.mean();} \par
(b) \texttt{data.mean(axis=0);} \par
(c) \texttt{np.mean(data);} \par
(d) \texttt{data.mean(axis=1);} \par
}

This question requires students to retrieve, from their memory, not only the correct function but also 
the appropriate parameter for handling dimensional directions, 
a common task in data science workflows. By regularly engaging in such 
low-stakes retrieval activities, students can reinforce their understanding of 
programming syntax and behavior, 
potentially leading to better learning outcomes.

\subsection{Challenges of Applying Retrieval Practice in Real-Time Teaching}

Despite the well-documented benefits of retrieval practice, 
there are significant challenges for instructors to implement them, in particular, during live teaching. 
A primary challenge is the inherently time-consuming and labor-intensive nature of generating 
high-quality, contextual, and appropriately challenging retrieval practice questions. 
This challenge is particularly evident in rapidly evolving technical fields such as 
Data Science, where course content is quickly updated, and the instructor needs to
come up with a set of new retrieval practice each time the course is taught.

Another challenge is to dynamically generate retrieval practices that 
achieve what is pedagogically known as "desirable difficulty". 
Effective retrieval practices not only recall factual knowledge, 
but also access higher-order thinking skills. During real-time teaching, 
it is necessary to adjust the level of difficulties based on students' performance.
If a previous activity is too easy, the subsequent practice should 
be made progressively more challenging, and vice versa.

\subsection{The Transformative Potential of Large Language Models (LLMs)}

Large language models (LLMs) such as OpenAI's GPT and Google's Gemini offer 
excellent capabilities in natural language understanding and generation.
These sophisticated models are engineered to process vast quantities of text data.
By analyzing various linguistic patterns, LLMs are able to generate coherent, 
contextually appropriate content. For educational purposes, LLMs hold significant
potential to automate a variety of tasks, for example, generating retrieval practice
questions based on relevant course materials. The automation capability can significantly reduce 
the burden on instructors, enabling them to focus more on course planning, delivering, 
and student engagement. 

\subsection{Contributions of This Study}

While the capabilities of LLMs in generating educational content are increasingly recognized, empirical evidence 
specifically examining the impact of LLM-generated retrieval practice questions on 
actual student learning outcomes within higher education, particularly in STEM fields, 
remains largely underexplored. 
Existing preliminary research presents mixed results, with many studies primarily focusing on 
the quality of question generation rather than the direct impacts on student 
learning \cite{olney2023generating,doughty2024comparative}. This indicates a critical need for rigorous investigation into the 
practical benefits of integrating LLMs into core teaching practices.   

In this study, we attempt to bridge the gap. We empirically investigate whether 
the administration of LLM-generated multiple-choice retrieval practice questions 
would enhance student learning outcomes. We conduct the research  
in college-level Data Science courses which cover data science programming in Python and cloud 
computing technology.

\noindent
\textbf{Summary and Contributions:} 
Each course lasted 10 weeks in a quarter system. We conducted the experiments in
weeks 6-9, a total of 4 weeks. Each experiment consisted of a 2-week cycle. 
In the first week of each cycle, students were either 
exposed to LLM-generated retrieval practice questions or received no 
retrieval practice. In the second week of each cycle, 
we assessed students' knowledge rentention with a set of quizzes.
The results clearly indicated that when students were 
exposed to LLM-generated retrieval practice, they achieved 
an average accuracy of 89\% in the assessment quiz. 
If students received no retrieval practice, they 
only achieved an averagy accuray of 73\%.

This study provides a piece of empirical evidence for the effectiveness of 
LLM-generated retrieval practice questions in a real-world higher education setting. 
It demonstrates a practical and scalable solution to the time-consuming of applying
an effective technique for real-time STEM teaching.

The remainder of the paper is organized as follows.
Section \ref{sec:related-work} reviews related work on using LLMs for 
multiple-choice question (MCQ) generation and evaluating their impact.
Section \ref{sec:methodology} outlines the study methodology and experimental design.
Section \ref{sec:results} presents the results of the study.
Section \ref{sec:discussion} discusses 
practical issues and limitations of the study.
Finally, Section \ref{sec:conclusion} discusses future directions and 
concludes the paper.

%
%
\section{Related Work}
\label{sec:related-work}
We review the related work in the following sub topics: 
\emph{the effectiveness of retrieval practice, 
automatic generation of MCQs using LLMs}, 
\emph{empirical studies on LLM-generated educational assessments},
and \emph{LLM application in broader
educational support and assessment}.

\subsection{The Effectiveness of Retrieval Practice}
The effectiveness of retrieval practice is highlighted by several key 
cognitive mechanisms, extensively studied in educational psychology 
\cite{roediger2011critical,Karpicke2007,Roediger2006-test-enhanced,Agarwal2008-examining,Nadel2012-memory,Roediger2011-test-enhanced,Lyle2011-retrieving,Agarwal2019-retrieval}.
This fundamental phenomenon illustrates that the active retrieval of information 
from memory significantly enhances long-term retention more effectively 
than passive re-studying. It is recognized not merely as an assessment tool but as 
a powerful teaching strategy \cite{powerful-teaching}. To achieve the desirable difficulties
in retrieving memory, retrieval practice questions should be designed with 
spacing \cite{Rohrer2006-effects,Maddox2016-understanding,Cepeda2008-spacing,Mcdaniel2011-test,Uner2017-effect}
and interleaving \cite{Rohrer2012-interleaving,Taylor2010-effects,Rohrer2015-interleaved,Roediger2014-forgetting,Pan2015-interleaving}.

\subsection{Automatic Generation of Multiple-Choice Questions (MCQs) using LLMs}

LLMs have been utilized to generate multiple-choice questions (MCQs) across a range of 
subjects. 
An empirical study by Olney \cite{olney2023generating} 
reported no significant differences between human-authored and 
LLM-authored MCQs when evaluated by medical experts for an anatomy 
and physiology textbook. This suggests a baseline level of quality can be achieved. 
Doughty et al. in \cite{doughty2024comparative} demonstrated that LLMs could 
produce high-quality programming MCQs aligning with 
learning objectives. 
Maity et al. in \cite{maity2024novel} studied a 
novel multi-stage promting (MSP) 
approach to generate zero-shot and one-shot MCQs in multiple human languages
(English, German, Hindi, Bengali) without fine-tuning. 
Tran et al in \cite{tran2023generating} found that GPT-4 significantly outperformed 
GPT-3 in generating isomorphic MCQs for computing courses, correctly producing 
answers for 78.5\% of questions compared to GPT-3's 30.8\%-36.7\%. 
While promising, some generated MCQs still contained multiple correct choices among distractors.
Lee et al. in \cite{lee2024few} generated English reading comprehension questions with 
high validity and reliability using ChatGPT. 
Biancini et al. in \cite{biancini2024multiple} also compared different LLMs for generating MCQs.
Maity et al. in \cite{maity2406effective} investigated
LLM's ability to generate questions align with different cognitive levels of Bloom’s Revised Taxonomy.

Addressing the generation of plausible and high-quality incorrect options 
(distractors) for MCQs is a particular challenge for assessment design. 
The survey in \cite{awalurahman2024automatic} indicated
that, despite advancements, challenges remained in ensuring plausibility, reliability, and 
diversity, with issues like hallucination and bias persisting. 
McNichos et al. in \cite{mcnichols2023automated} showed that in-context learning with selected 
examples was more effective than random or zero-shot baselines for math MCQs.
Bitew et al. in \cite{bitew2023distractor} demonstrated that ChatGPT-driven solutions 
could produce high-quality distractors and were more reliable, generating 
significantly fewer "nonsense distractors" compared to traditional ranking-based models.

Extensions to theses studies include 
developing advanced LLM techniques to generate distractors that 
accurately reflect common student misconceptions rather 
than just arbitrary incorrect options, 
creating more sophisticated reference-free metrics for evaluating distractor 
quality to better 
capture their pedagogical effectiveness, further exploring the automated generation 
of informative feedback messages for each distractor, and explaining why
an option is incorrect and guiding students toward the correct understanding.

\subsection{Empirical Studies on LLM-Generated Educational Assessments}

The empirical investigation in \cite{witsken2025llms} by Witsken et al.  
found that while students could not discern whether questions were LLM-authored or 
human-authored, student scores on LLM-authored questions were almost 9\% lower. 
This outcome could indicate subtle differences in difficulty or styles that might 
impact student performance, even if the questions are deemed ``well-composed and relevant" by 
human reviewers. 

Another empirical study in \cite{yusof2025chatgpt} by Yusof compared 
the final exam scores between previous cohorts who did not use ChatGPT for 
retrieval practice (control group) and current cohorts who did (experimental group). 
The study reported significantly better final exam scores for students in the experimental group.
The primary sample consisted of second-year education students enrolled in the Measurement 
and Evaluation in Education course, with 64 students randomly selected for each group. 
The study also highlighted the need of human monitoring and judgment 
for addressing complex learning needs and provide deeper contextual understanding. 

\subsection{LLM Application in Broader Educational Support and Assessment}

LLMs are also applied to a wider array of educational tasks beyond the direct 
generation of questions, including pedagogical support, content creation, 
and general educational research. 
Hu et al. in \cite{hu2025exploring} demonstrated that LLMs (GPT-4, Claude 3.5 Sonnet) could significantly 
enhance the quality of teaching plans. This was achieved through a multi-stage 
process involving simulating teacher-student interactions, generating 
teaching reflections, and iteratively refining the original teaching 
plan based on these simulated insights. 
Das et al. in \cite{das2021automatic} surveyed broadly on automatic question generation and 
answer assessment in online learning. It highlighted challenges in assessing subjective 
questions and the need for standardized evaluation procedures. 

Despite these developments and advancements in various educational aspects, 
there remains a need for more empirical studies, 
particularly in STEM fields, to rigorously assess the effectiveness of 
LLM-generated content for retrieval practice activities.   

%
%
\section{Methodology}
\label{sec:methodology}

Our investigation is motivated by the following research question: 
\emph{Does frequent use of LLM-generated MCQs as retrieval practice during 
real-time STEM instruction improve student knowledge retention?} To answer this 
question, we implement a quasi-experiment within real classroom environments.

\subsection{Study Design: A Quasi-Experimental Approach}
While true experiments, characterized by random assignment, are ideal for 
evaluating the effect of the intervention, i.e., LLM-generated MCQs, 
their implementation is impractical in real-world classrooms.
It is difficult to randomly assign students of the same live lecture to different
intervention groups without introducing performance bias due to lack of blinding. 
A quasi-experimental design, by contrast, enhances the practical relevance 
of the study's conclusions for real-world and authentic context.

Specifically, we utilize a non-equivalent groups design, wherein pre-existing 
student groups were compared, rather than relying on random individual assignment. 
Students within the same course assigned to different 
conditions in different sections. In one section, students 
receives the intervention (e.g., LLM-generated MCQs as retrieval practice), and 
the other does not. We 
compare the outcomes (e.g., knowledge retention) over time to infer the 
intervention’s effect.

However, it is important to acknowledge the inherent limitations of quasi-experimental designs 
regarding internal validity when compared to true experiments. 
The absence of true random assignment means that unmeasured confounding 
variables could potentially influence the observed effects, as 
pre-existing differences between sections cannot be entirely ruled out. 

\subsection{Participants and Context}

The study cohort comprised approximately 60 undergraduate students enrolled 
in two college-level courses in an undergraduate data science program. 
Both courses address rapidly evolving technical domains, 
requiring students to develop strong conceptual understanding and 
long-term retention in these domains.
Each course lasted 10 weeks in a quarter system.

\subsection{Implementation of the Study}

The empirical investigation was conducted over a total of four weeks, 
specifically weeks 6-9 of the 10-week quarter system for both courses, 
named as \textbf{course1} and \textbf{course2}. 
The experiment was structured into two distinct 2-week cycles.

\subsubsection{First Cycle (Weeks 6-7): LLM-Enhanced Retrieval Practice Condition}

\noindent
\textbf{Week 1 of First Cycle (Week 6 in the quarter):} The students in \textbf{course1} and 
\textbf{course2} were taught using their respective content. Each course included 
two lectures, each lasting 120 minutes. The instructor deliberately divided 
each lecture into 10–15-minute segments. At the end of each segment, students 
were asked to answer multiple-choice questions (MCQs) generated by an LLM as a form of retrieval practice.
Immediate feedback on the correctness of responses was provided to students after each attempt. 
The retrieval practice activities were explicitly designed as low-stakes learning opportunities.
Students could try the retrieval practice as many times as they want.

\noindent
\textbf{Week 2 of First Cycle (Week 7 in the quarter):} 
The students in both courses took a comprehensive quiz (Quiz 1), respectively.
The quiz covered the material presented in Week 1 of this cycle.

\subsubsection{Second Cycle (Weeks 8-9): Control Condition (No Retrieval Practice)}

\noindent
\textbf{Week 1 of Second Cycle (Week 8 in the quarter):} 
The students in \textbf{course1} and 
\textbf{course2} were taught with their respective content across both lectures. 
They were allowed to review the lecture materials and re-study the content 
independently. However, no retrieval practice activities were 
incorporated during the live lectures.

\noindent
\textbf{Week 2 of Second Cycle (Week 9 in the quarter):} 
The students in both courses took another 
comprehensive quiz (Quiz 2) covering the material presented in Week 
1 of this second cycle.

The online platform used for the  delivery of MCQs and quizzes was 
Google form.

\subsection{Generating MCQs Using LLM}

Our experiment was conducted from March to June 2025. During this 
period, we accessed the free versions of ChatGPT (GPT-3.5 and GPT-4o) 
and Google AI Studio (Gemini 2.0 and 2.5 Flash). After initial 
comparisons of MCQs generated from the same lecture materials across 
multiple rounds, we did not observe significant differences between 
the applications. Consequently, we primarily used Google AI Studio 
for MCQ generation in the experiments.

\noindent
\textbf{Course Materials:} The teaching of both courses was delivered 
through Jupyter notebooks. We strategically placed breaks roughly every 10-15 
minutes within the notebooks for retrieval practice activities. 
For the purpose of generating MCQs, each notebook's content was converted 
into markdown text, and the relevant sections were injected 
into the AI prompts.

\noindent
\textbf{Prompt Engineering:} To ensure the generation of high-quality, contextually 
relevant, and pedagogically sound multiple choice questions (MCQs), we iteratively 
crafted the following prompt:

\begin{verbatim}
    ### System Instruction ###
    You are an expert college course instructor teaching data science 
    programing and cloud computing. Your task is to create high-quality, 
    pedagogically sound multiple choice questions (MCQs) based on the 
    course content provided below. Create these MCQs that test:

    1. Factual recall: checking students' understanding of specific 
       concepts or facts.
    2. Higher-order thinking: such as application, analysis, or 
       evaluation.

    Each MCQ must:
    - Include one correct answer and three plausible distractors.
    - Avoid using clues or phrasing in the question stem that would 
      directly give away the correct answer.
    - Ensure that all answer options are grammatically parallel, 
      plausible, and contextually relevant.
    - Mix the cognitive level across questions: some should be direct, 
      others should require critical thinking.

    Output Each MCQ in the Following Format:
        Question: [Insert question here]

        Options (One Correct Answer):
            (a) Option A  
            (b) Option B  
            (c) Option C  
            (d) Option D

        Correct Answer: [A/B/C/D]  
        Explanation: [Brief explanation of why it's correct, and why 
        distractors are incorrect.]

    ### User Instruction ###
    Generate at least 15 MCQs based on the following course content.
    """ 
    {Insert course content here}
    """

\end{verbatim}

The generated MCQs will undergo the following quality assurance checks: 

\begin{itemize}
    \item Ensuring questions accurately reflected the provided lecture content and 
 aligned with learning objectives.   

    \item Verifying that questions and their respective options were clearly 
    formulated and unambiguous.   

    \item Confirming that the designated correct answer was indeed accurate and sound.   

    \item Assessing that incorrect options (distractors) were plausible and challenging enough 
    to test understanding, preventing easy identification of the correct answer.   

    \item Ensuring questions assessed a range of cognitive levels, from basic factual recall 
    to high-order thinking   

    \item Verifying that questions contained no hallucinations.   
 
\end{itemize}

As we will discuss in Section \ref{sec:discussion}, 
while LLMs offer significant scalability, sometimes they still
produce low-quality MCQs. Currently, human-in-the-loop validation 
is a critical component before the instructor release the MCQs to students.   

\subsection{Data Collection and Statistical Analysis}

\noindent
\textbf{Data Points:} For each course, data were collected during the second 
week of each cycle through quizzes. Quiz 1, administered in the first cycle, was 
conducted under the LLM retrieval practice condition, whereas Quiz 2, administered in 
the second cycle, served as the control condition with no LLM-generated 
MCQs used for retrieval practice. The primary data collected for analysis 
were student quiz scores, measured as the percentage of correct answers 
for each participant on both Quiz 1 and Quiz 2. We then combined the results of 
Quiz 1 from both courses into a single list. Similarly, the results of Quiz 2 from 
both courses were merged into another list.

\noindent
\textbf{Statistical Analysis:} 
The lengths of the two datasets differ due to variations in course attendance
and student response. However, we ensured that most of the students remained the 
same, in particular, for cycle 1.
In addition, the quizzes were administered anonymously, so the participants 
could be treated as independent groups. An appropriate statistical test will be 
conducted to determine whether there was a statistically significant 
difference in the results between Quiz 1 and Quiz 2.

\begin{itemize}
    \item \emph{Assumptions:} Prior to conducting the test, we will check the 
distributions of the difference scores between the two quizzes.   
A p-value of $\leq 0.05$ is predetermined as the threshold for statistical significance. 

    \item \emph{Effect Size:} 
In addition to statistical significance, effect size will be 
calculated to quantify the practical magnitude of the observed difference in scores. 
This measure provides a clearer understanding of how large the difference between 
the quiz scores truly is, independent of sample size. Effect size would illustrate 
the real-world impact rather than just the likelihood of the difference occurring by chance.
\end{itemize}

%
%
\section{Results}
\label{sec:results}

\begin{figure}[h!]
    \centering
    \includegraphics[width=0.9\textwidth]{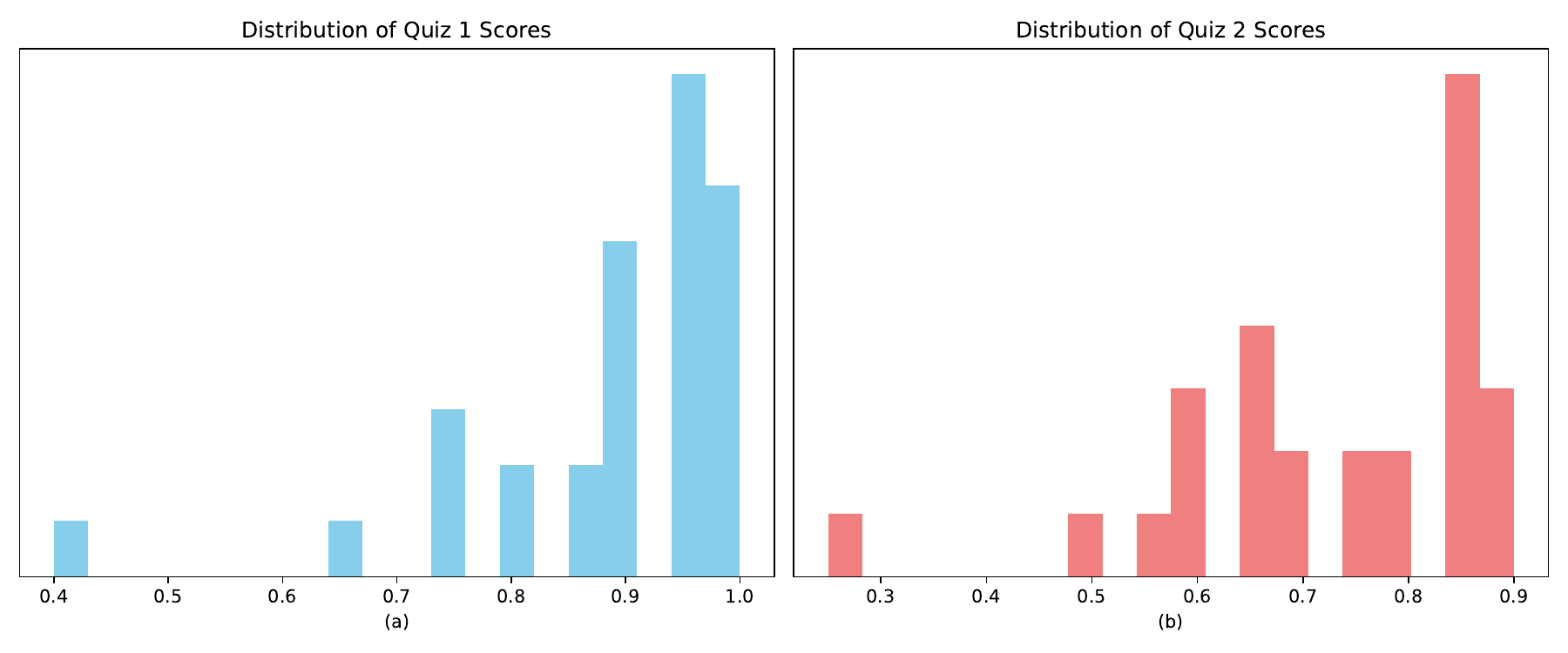}
    \caption{Score Distributions of Quiz 1 and Quiz 2}
    \label{fig:score-distribution}
\end{figure}

Figure \ref{fig:score-distribution} (a) and Figure \ref{fig:score-distribution} (b) 
show the score distributions of Quiz 1 and Quiz 2, respectively. 
The visualization indicates that the scores are skewed and not normally distributed. 
Also, the two lists have different lengths. By these observations, we conduct Mann-Whitney U test which 
is non-parametric and compares the medians. A common effect size measure for 
the Mann-Whitney U test is $r$ or rank-biserial correlation. 
This effect size measures the proportion of times a randomly selected observation 
from one list is larger than a randomly selected observation from the 
other list, minus the reverse proportion.

In essence, 
\begin{itemize}
    \item The Mann-Whitney U test answers the question: \texttt{"Is there a statistically significant 
tendency for scores in one quiz to be higher than scores in the other?"}

    \item The rank-biserial correlation $r$ answers the question: \texttt{"How large is this tendency or 
difference in distributions?"} A larger r (between -1 and 1) value indicates a greater degree of 
separation or difference between the distributions of scores from the two quizzes.
\end{itemize}

Table \ref{tab:test-results} shows the results of the Mann-Whitney U test. 
The value of U statistics is 703 which is closer to $n1 * n2=837$. The larger value 
suggests less overlap in the ranks, indicating a greater separation between the two groups.
The small p-value indicates that we have sufficient evidence to conclude that there 
is a statistically significant difference in the distributions of scores between the two quizzes.

This finding, combined with the large positive Z-score (4.4627), the large positive 
rank-biserial correlation (r = 0.5860), and the descriptive statistics showing a much higher 
median and mean for Quiz 1, provides very strong evidence that students performed 
significantly better on Quiz 1 than on Quiz 2. The observed difference is 
highly unlikely to be due to random chance.

\begin{table}[!ht]
    \begin{center}
	\begin{tabular}{l | l}
	\hline
         \textbf{Item} & \textbf{Results} \\
         \hline
         The Mann-Whitney U test: & \texttt{U Statistic: 703.0000} \\
                                  & \texttt{P-value < 0.0001 }\\
        \hline
         Effect Size: & \texttt{Sample size Quiz 1 (n1): 31} \\
                     & \texttt{Sample size Quiz 2 (n2): 27} \\
                     & \texttt{Total sample size (N): 58} \\
                     & \texttt{Approximate Z-score: 4.4627} \\
                     & \texttt{Rank-Biserial Correlation (r): 0.5860} \\
        \hline    
        Quiz 1 Scores: & \texttt{mean: 0.89} \\
                       & \texttt{std: 0.13} \\
                       & \texttt{median: 0.95} \\
        \hline
        Quiz 2 Scores: & \texttt{mean: 0.73} \\
                       & \texttt{std: 0.15} \\
                       & \texttt{median: 0.75} \\
        \hline
        \end{tabular}
    \end{center}
    \caption{The Results of Statistical Analysis}
    \label{tab:test-results}
\end{table}

%
%
\section{Discussion}
\label{sec:discussion}

The statistical analysis on scores from Quiz 1 and Quiz 2 revealed that the 
students achieved significantly better learning outcomes if they practiced with 
LLM-generated MCQs, compared to a time when they did not perform 
any retrieval practice.  Although LLMs present 
a scalable solution for instructors  to incorporate retrieval 
practice activities into live teaching, some practical considerations exist.
In this section, we discuss the common quality issues of the LLM-generate MCQs
and the limitations of the study.

\subsection{Quality Issues of the LLM-Generated MCQs}

We have been using LLMs to generate various assessment materials since ChatGPT appeared 
in Nov. 2022. Our experience has shown that the quality of LLM-generated content 
varies significantly depending on LLM, prompt design, and knowledge injection. Human 
verification and validation is indispensable. For example, we always presented 5 MCQs as retrieval 
practice for every 10-15-minute lecture. To choose 5 high-quality MCQs, we typically had to ask 
the LLM to generate at least 15 MCQs based on given course content. There were about 2/3 of the 
LLM-generated MCQs which did not meet our objectives. These low-quality MCQs can be categorized
into following issues: \emph{(1) question and options containing mistakes, (2) hallucinations, 
(3) question containing hints to answers, (4) duplicated options, (5) ambiguous answers, 
(6) easy distractors, (7) trivial knowledge, (8) inconsistent formats of options, (9) only negations
in distractors}, \emph{(10) double-concept questions, (11) incorrect assumption}, 
and , \emph{(12) ambiguous question}. In the following, we 
illustrate each issue using an example we have encountered. Note that the MCQ examples below
were generated by an LLM based on the course materials injected in the prompt.

\noindent
\textbf{1. Question and options containing mistakes.} LLMs generate incorrect MCQs from time to time. 
For example, when we extract a subset of a 2-D numpy array, we can use either positive indices counting from 
0 or negative indices counting from -1 backwardly. The LLM struggled in  
reasoning about negative indices. Here is an example with mistakes:
\begin{verbatim}
    Question: "What is the result of slicing arr[-3:, :-1] 
    where arr = np.array([[1, 2, 3], [4, 5, 6], [7, 8, 9]])?"

    Options (One Correct Answer):
        (a) [[7, 8], [4, 5]]
        (b) [[1, 2], [7, 8]]
        (c) [[4, 5], [7, 8]]
        (d) [[7, 8], [4, 5], [1, 2]]
\end{verbatim}
Based on the course materials we taught in the lecture, 
the correct answer is \texttt{[[1, 2], [4, 5], [7, 8]]}. But after we injected the 
course materials to the LLM, the options generated by 
the LLM do not contain the correct answer. The LLM reasoned about the content incorrectly.

\noindent
\textbf{2. Hallucinations.} LLMs may generate question and options that contain concepts
which do not exist in the course materials. For example, the following MCQ contains
the attribute \texttt{arr.rank}. However, the lectures never mentioned the 
attribute \texttt{arr.rank}. The LLM made up such a concept in the MCQ question.
\begin{verbatim}
    Question: "The arr.rank attribute returns the number of axes 
    in a NumPy array."

    Options (One Correct Answer):
        (a) True
        (b) False
        (c) Only if arr.shape is not defined
        (d) Only in NumPy v2.0+
\end{verbatim}

\noindent
\textbf{3. Question containing hints to answers.}
LLMs may generate a question that contains a hint to the correct answer. For example, 
in the following MCQ, the hint 'What is Spark' is contained in the question.
\begin{verbatim}
    Question: "What is the name of the unified analytics engine for 
    big data processing that is mentioned in the `What is Spark?' section?"
    
    Options (One Correct Answer):
        (a) Hadoop
        (b) Spark
        (c) Flink
        (d) MapReduce
\end{verbatim}

\noindent
\textbf{4. Duplicated options.} LLMs may generate MCQs that have duplicated options. For example
in the following MCQ,  the options (c) and (d) are duplicated because
the words ``mean" and ``average" have the same meaning.
\begin{verbatim}
    Quesiton: "What does np.mean(arr, axis=1) compute when 
    arr is a 2D array?"

    Options (One Correct Answer):    
        (a) The mean of all elements
        (b) The mean of each row
        (c) The mean of each column
        (d) The average of each column
\end{verbatim}

\noindent
\textbf{5. Ambiguous answers.} LLMs may generate MCQs that have ambiguous answers.
For example, in the following MCQ, the options (c) and (d) are ambiguous. 
Either one could be a correct answer given some reasonable explanation.
\begin{verbatim}
    Question: "What is the cost function used for training logistic 
    regression models?"

    Options (One Correct Answer):
        (a) Sigmoid function 
        (b) Gradient function 
        (c) Log likelihood function 
        (d) Cross-entropy function. 
\end{verbatim}

\noindent
\textbf{6. Easy distractors.} LLMs may generate MCQs that provide one correct
option and three easy distractors. For example, the distractors of the
following MCQs can be discerned without much knowledge about the content.
\begin{verbatim}
    Question: "What does arr.reshape(2, 3) do to a 1D array of 6 elements?"

    Options (One Correct Answer):
        (a) Changes its shape to 2 rows and 3 columns
        (b) Removes the last 3 elements
        (c) Sorts the array
        (d) Adds padding to reach a 2x3 shape
\end{verbatim}

\noindent
\textbf{7. Trivial knowledge.} LLMs may generate MCQs that are trivial for factual recall.
For example, the following MCQ asks a trivial question. 
\begin{verbatim}
    Question: "Which property returns the shape of a NumPy array?"

    Options (One Correct Answer):
        (a) ndim
        (b) size
        (c) shape
        (d) dtype
\end{verbatim}

\noindent
\textbf{8. Inconsistent formats of options.} LLMs may generate MCQs with 
inconsistently formatted options that make the correct answer easily 
identifiable. For example, in the following MCQ, the correct option is 
significantly longer than the distractors, making it stand out.
\begin{verbatim}
    Questions: "What is the primary reason that single machines can 
    struggle with large-scale data processing?"

    Options (One Correct Answer):
        (a) Lack of available memory
        (b) Slow network connection
        (c) Not enough hard drive space
        (d) Single machines often lack the sufficient processing power 
            and resources, as well as the speed required, to efficiently 
            perform computations on the vast amounts of information that 
            characterize big data, leading to prohibitively long 
            computation times.

\end{verbatim}

\noindent
\textbf{9. Only negations in distractors.} LLMs may generate MCQs in which 
all distractor options contain negations, while the correct answer 
does not. For example, the following MCQ has such a pattern that 
makes it easy identify the correct answer simply by recognizing 
the difference in phrasing, rather than understanding the content.

\begin{verbatim}
    Question: "What is true about sequential feature selection algorithms?"

    Options (One Correct Answer):
        (a) They reduce the feature space step-by-step using a 
            performance criterion
        (b) They do not consider classification performance while 
            reducing features
        (c) They avoid using greedy search methods
        (d) They never remove features from the original set

\end{verbatim}

\noindent
\textbf{10. Double-concept questions.} LLMs may generate MCQs that 
contain multiple concepts in the description causing confusion. For example, 
the following MCQ describes disconnected parts in the question with insufficient
context.

\begin{verbatim}
    Question: "According to the text, overfitting is characterized by high 
    variance and can be addressed by regularization. However, if a model is 
    suffering from underfitting , what action would generally be more 
    appropriate based on the principles discussed?
    
    Options (One Correct Answer):
        (a) Reduce regularization
        (b) Increase regularization
        (c) Add more data
        (d) Add more features 

\end{verbatim}

\noindent
\textbf{11. Incorrect Assumption.} LLMs may generate MCQs by making incorrect assumptions. For example,
in the following MCQ, the LLM assumed that the order of steps mentioned in the lecture was 
the only one that is correct. However, the order was just suggested by the API.
The LLM assumed that the suggested one was the only one by using the phrase "the correct order of steps."
\begin{verbatim}
    Question: "According to the Scikit-Learn estimator API, which of the 
    following is the correct order of steps?"

    Options (One Correct Answer):
        (a) 1. Fit the model, 2. Choose model, 3. Arrange data, 
            4. Apply the Model
        (b) 1. Choose model, 2. Choose hyperparameters, 3. Arrange data, 
            4. Fit the model, 5. Apply the Model
        (c) 1. Arrange data, 2. Choose model, 3. Choose hyperparameters, 
            4. Fit the model, 5. Apply the Model
        (d) 1. Arrange data, 2. Fit the model, 3. Apply the Model, 
            4. Choose model, 5. Choose hyperparameters

\end{verbatim}

\noindent
\textbf{12. Ambiguous Question.} LLMs may generated MCQs that have ambiguous questions. 
For example, the following question does not refer to the context in the course materials.
Additionally, the term 'feature' is ambiguous since it could refer to names or data.
\begin{verbatim}
    Question: "What data structure is most often used to contain 
    the features?"

    Options (One Correct Answer):
        (a) Pandas Series
        (b) Python List
        (c) NumPy array or Pandas DataFrame
        (d) Python Dictionary
\end{verbatim}

\subsection{Limitations of the Current Study}

Although the results support the hypothesis that LLM-generated retrieval 
practice questions significantly enhance student learning outcomes, 
there are several limitations in the current study. 
The first limitation is the quasi-experimental design. 
Confounding variables such as the differences between 
course materials in different sections 
could potentially influence the observed results. 
Secondly, the study involved a relatively small sample size. 
The generalizability of these findings to other academic disciplines, 
larger student populations, or different educational levels (e.g., K-12) 
may therefore be limited.

Furthermore, the assessment was conducted in 2-week cycles, measuring relatively 
short-term knowledge retention. We also used exclusively 
multiple-choice questions (MCQs) for retrieval practice. Other types of questions
such as short-answer questions, or more complex problem-solving tasks could have deeper impacts.
In addition, we did not measure student perceptions of the LLM-generated MCQs or 
their overall learning experience with the intervention. Finally, 
we did not measure the time and effort required for human validation. 

Some ethical considerations include responsible AI integration into 
classroom, instructor training, bias reduction, and maintaining 
the development of critical-thinking
and complex problem-solving skills.

%
%
\section{Conclusion}
\label{sec:conclusion}

This empirical study provides a piece of evidence demonstrating that integrating 
LLM-generated retrieval practice questions into higher education STEM courses 
significantly improves student learning outcomes. Students exposed to 
LLM-generated multiple-choice questions (MCQs) achieved a 16-percentage-point 
higher quiz accuracy (89\% vs. 73\%) compared to a control group. 
This observed difference was not only statistically significant but 
also practically meaningful, indicating a substantial positive impact of LLMs 
on learning. 

The study encourages us to actively explore and develop an AI-powered 
lecture assistant (AILA) system \cite{an2024lecture} that can automatically detect a favorite 
moment for intervention. Future research directions include (1) automatic
quality control of the LLM-generated retrieval practice questions, 
(2) studying the impacts of such a AI-powered lecture assistant (AILA) system on students
long-term retention in diverse teaching environments, and 
(3) investigating the instructor's perceptions and engagement in using 
such a AI-powered lecture assistant (AILA) system during course preparation and 
live teaching.

\newpage
%
%
\bibliographystyle{acm} 

\end{document}